\documentclass{article}
\usepackage{spconf, amsmath, graphicx}
\usepackage{algorithm}
\usepackage{algorithmic}
\usepackage{multirow}
\usepackage{booktabs}
\usepackage{amssymb}

\title{OPT: One-shot Pose-Controllable Talking Head Generation}
\name{Jin Liu$^{1,2}$\thanks{$^{\ast}$Corresponding authors. \newline This research is supported in part by  the National Key Research and Development Program of China (No. 2020AAA0140000), and the National Natural Science Foundation of China (No. 61702502).}, Xi Wang$^{1*}$,  Xiaomeng Fu$^{1,2}$, Yesheng Chai$^{1}$,Cai Yu$^{1,2}$,Jiao Dai$^{1*}$,Jizhong Han$^{1}$}

\address{$^{1}$Institute of Information Engineering, Chinese Academy of Sciences, Beijing, China\\
	$^{2}$School of Cyber Security, University of Chinese Academy of Sciences, Beijing, China\\}
	
\begin{document}
%
\maketitle
\begin{abstract}
One-shot talking head generation produces lip-sync talking heads based on arbitrary audio and one source face.  To guarantee the naturalness and realness, recent methods propose to achieve free pose control instead of simply editing mouth areas. However, existing methods do not preserve accurate identity of source face when generating head motions. To solve the identity mismatch problem and achieve high-quality free pose control, we present One-shot Pose-controllable Talking head generation network (OPT). Specifically, the Audio Feature Disentanglement Module separates content features from audios, eliminating the influence of speaker-specific information contained in arbitrary driving audios.  Later, the mouth expression feature is extracted from the content feature and source face, during which the landmark loss is designed to enhance the accuracy of facial structure and identity preserving quality. Finally, to achieve free pose control, controllable head pose features from reference videos are fed into the Video Generator along with the expression feature and source face to generate new talking heads. Extensive quantitative and qualitative experimental results verify that OPT generates high-quality  pose-controllable talking heads with no identity mismatch problem, outperforming previous SOTA methods. 
\end{abstract}


\section{Introduction}
Talking head generation aims to drive the source face image with the audio signal and produces a lip-sync talking head video, which is significant to various practical multimedia applications, such as film making, virtual education, video  conferencing,  digital human animation and short video creation.

Talking head generation can be divided into two categories: speaker-specific methods and speaker-independent methods. The speaker-specific methods~\cite{zhang2021facial, guo2021ad, zhang2022meta} only generate talking heads of fixed subject and requires large amount of person-specific high-quality videos, which limits the application and generalization.

The speaker-independent methods are designed to animate video portraits given one unseen source face and driving audio. Some one-shot works~\cite{chen2019hierarchical, prajwal2020lip} simply edit the mouth region and keep the other areas of source face unchanged. Their generated talking head videos are unnatural with the fixed facial contour, blending traces around the mouth and no head motion changes. Therefore, current one-shot speaker-independent works focus on full-frame generation~\cite{zhou2021pose,wang2021audio2head, liang2022expressive}, which produce the whole head areas, together with neck parts and background.  

To improve the naturalness and realness, some methods propose to add natural head poses into talking heads. PC-AVS~\cite{zhou2021pose} modularizes audio-visual representations by devising an implicit low-dimension pose code.  Audio2Head~\cite{wang2021audio2head} utilizes a motion-aware recurrent neural network to predict head motions from audio. However, in talking heads of above methods with new poses, the source identity is not well preserved due to the facial structure change, as shown in Fig. \ref{fig:quality}.

The identity mismatch problem means the inability of generated talking heads to preserve the identity of source faces.  Previous image driven face reenactment works~\cite{liu2021li} focus on solving the identity mismatch problem in visual modality, which caused by the inconsistent facial contour between driving subject and source person. When it comes to audio-driven paradigm, the gap between audio and visual modality becomes even larger. All the  information contained in audio signal affects the driving representation extraction, among which the content feature is the most important since it directly relates to the mouth shape. Given the fact that different speakers' audios with the same content are different, we believe that it is necessary to disentangle identity and content features from audio signal.  Furthermore, it is important to extract accurate mouth expression features since the facial structure changes when performing different head poses.

Specifically, we present the \textbf{O}ne-shot \textbf{P}ose-controllable \textbf{T}alking head generation network (OPT). The \textit{Audio Feature Disentanglement Module} separates identity and content features explicitly from audio signals. Later, the facial expression feature is extracted from content feature and source face, during which the landmark loss is designed to enhance the accuracy of facial structure and identity preserving quality. Finally, the head pose feature reconstructed from other pose videos using 3DMM~\cite{deng2019accurate}  is fed into the Video Generator along with expression feature and source face to generate new talking heads.
Extensive experimental results demonstrate the superior performance of OPT and the effectiveness of several modules. Our contributions are summarized as follows: 
\begin{itemize}
	\item The proposed OPT is the first to simultaneously perform one-shot identity-independent pose-controllable talking head generation with almost no \textit{identity mismatch problem}.
	\item To solve the identity mismatch problem, the \textit{Audio Feature Disentanglement Module} is proposed to successfully decompose intrinsic identity features and content features over audio signals. 
	\item The landmark loss is designed to enhance the accuracy of facial shape and the identity preserving quality.  The explicit head pose feature is also utilized to guide the free pose control.
	
\end{itemize}

\section{Our Method}
\begin{figure}[t]
	\centering
	\includegraphics[width=\columnwidth]{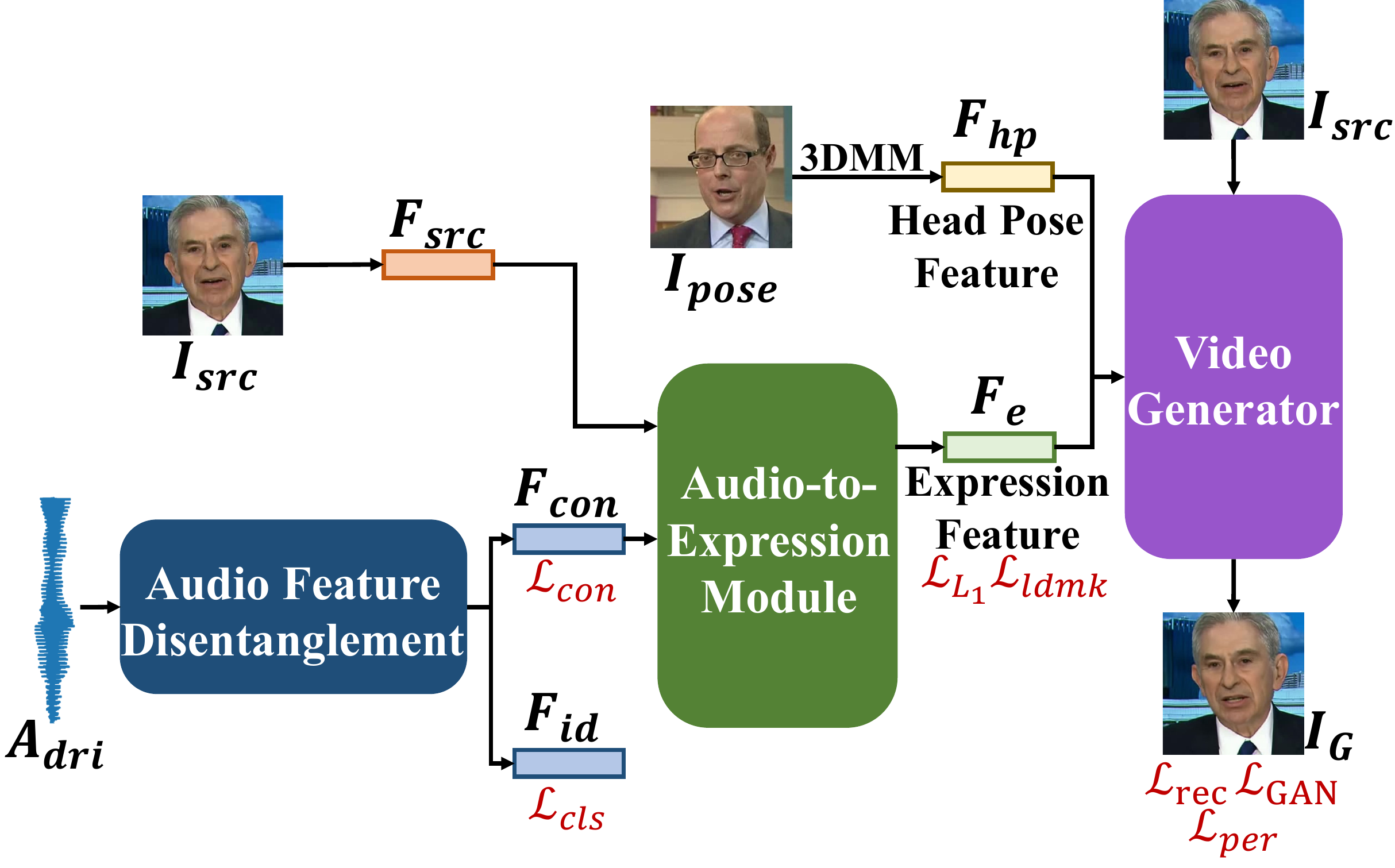} 
		\vspace{-6mm}
	\caption{{Overview of OPT.}  The Audio Feature Disentanglement Module separates content feature from driving audio. Then the Audio-to-Expression Module extracts the expression feature given conent feature and soure face. Finally, the head pose feature from other video and expression feature are fed into the Video Generator to generate new talking heads.   }
	\label{fig:overview}
\end{figure}

Fig. \ref{fig:overview} summarizes the pipeline of our proposed method. 
OPT takes driving audio $A_{dri}$, source image $I_{src}$ and pose image $I_{pose}$ as inputs to generate $I_{G}$, indicating $I_{src}$ speaking the corpus of $A_{dri}$ with the head pose of  $I_{pose}$. 
The \emph{Audio Feature Disentanglement Module} separates content feature $F_{con}$ and identity feature  $F_{id}$ from $A_{dri}$. 
Then $F_{con}$ and source feature $F_{src}$ from $I_{src}$ are fed into the \emph{Audio-to-Expression Module} to produce expression feature $F_{e}$. 
Finally, the \emph{Video Generator} takes $I_{src}$, $F_{e}$ and head pose feature $F_{hp}$ from $I_{pose}$ as inputs to generate $I_{G}$. Each module will be introduced detailedly in the following sections.

\subsection{Audio Feature Disentanglement Module}

Given that audios of the same content but of different speakers are diverse, we believe the inherently entangled identity and content features need to be independently extracted from audio signals, to achieve audio-based identity control for talking head generation. During inference, the identity of  $A_{dri}$ and $I_{src}$ are usually different, since the driving audio is chosen arbitrarily. Unlike previous methods~\cite{prajwal2020lip,zhou2021pose} who merely extract entangled features from audio signals, we propose the Audio Feature Disentanglement Module (AFDM)  to map audios into two separate latent audio spaces: a content-agnostic encoding space of the identity and a content-dependent encoding space of the corpus corresponding to audio. 

Specifically, AFDM contains two encoders $E_{con}$ and $E_{id}$ to individually extracts corresponding features from  $A_{dri}$. The content loss $	\mathcal{L}_{con}$ is as follows:

\vspace{-2mm}
\begin{equation}
		\mathcal{L}_{con} =  \left\| E_{con}\left( A \right)  -  E_{con}( \tilde{A} ) \right\| _{1}, 
		\label{distanglement_content}
		\vspace{-2mm}
\end{equation}
where audio $A$ and $\tilde{A}$ shares the same content but spoken by different subjects. The identity loss $\mathcal{L}_{id}$ is used to train $E_{id}$:

\vspace{-2mm}
\begin{equation}
	\mathcal{L}_{cls} =  - \sum_{i=1}^{N}
	\left( p * \log q \right),
	\label{distanglement_identity}
	\vspace{-2mm}
\end{equation}
where $N$ denotes the total amount of speaker identities, $p$ means the ground truth identity  probability and $q$ represents the $E_{id}$ prediction probability.  In this way, the pure content feature $F_{con}$ can be accurately separated from $A_{dri}$ through AFDM to guide the following extraction process of expression feature.

\subsection{Audio-to-Expression Module}
The Audio-to-Expression Module (AEM) predicts expression feature $F_{e}$ from features $F_{con}$ and $F_{src}$.
Considering the good reconstruction performance of 3DMM~\cite{deng2019accurate}, we use part of 3DMM parameters to represent the expression feature.
With a 3DMM, the 3D shape $\mathbf{S}$ of a face is parameterized as follows:

\vspace{-2mm}
\begin{equation}
\mathbf{S}=\overline{\mathbf{S}}+\boldsymbol{\alpha} \mathbf{B}_{i d}+\boldsymbol{\beta} \mathbf{B}_{e x p},
	\vspace{-2mm}
\end{equation}

where $\overline{\mathbf{S}}$ is the average face shape, $\mathbf{B}_{i d}$ and $\mathbf{B}_{e x p}$ are basis of identity and expression via PCA. The coefficients $\alpha \in \mathbb{R}^{80}$ and $\beta \in \mathbb{R}^{64}$ describe the facial shape and expression respectively. In our method, we choose $\beta$ as expression features $f_e$. The AEM contains multiple convolutional layers and linear layers. The $L1$ loss $\mathcal{L}_{L1}$ is imposed on $\tilde{F}_e$ and the ground truth $F_{e}$.  In order to further improve the lip-sync quality and keep the accurate facial contour structure to alleviate the identity mismatch problem,  we further design the facial landmark loss $\mathcal{L}_{ldmk}$ using 3DMM:

\vspace{-2mm}
\begin{equation}
	\mathcal{L}_{ldmk}=\frac{1}{N} \sum_{n=1}^{N} \omega_{n}\left\|\tilde{\mathbf{l}}_{n}-\mathbf{l}_{n}\right\|^{2}
	\vspace{-2mm}
\end{equation}
where $\{\mathbf{l}_{n}\}$ is facial landmarks of ground truth driving face image, $\{\tilde{\mathbf{l}}_{n}\}$ is the 3D landmark vertices projection of reconstructed shape using $f_e$ onto the image plane. $N$ denotes the number of landmarks. The weight $\omega_{n}$ id set to 20 for inner mouth and nose points and others to 1.

\subsection{Video Generator}
To achieve pose-controllable talking head generation, the additional driving pose face image $I_{pose}$ is need to provide head pose feature. During inference,  $I_{pose}$ could come from real talking head videos to offer auxiliary pose feature sequences.  The head pose feature $F_{hp}$ is denoted by rotation $R \in SO(3)$ and translation $T \in \mathbb{R}^{3}$, which could also be obtained during the process of 3DMM reconstruction~\cite{deng2019accurate}. Finally,  given $I_{src}$, $F_{hp}$ and $F_{e}$, the Video Generator(VG) produces new talking head$I_G$.  Detailedly, VG contains Conv and TransConv layers with residual connection. The two features are fed by AdaIn~\cite{huang2017arbitrary} opeation. The discriminator utilized PatchGAN~\cite{isola2017image} and VG minimizes the reconstruction loss $\mathcal{L}_{rec}$ between ground truth image $I$ and generated head image $I_G$.
\vspace{-2mm}
\begin{equation}
	\mathcal{L}_{rec}=\|I-I_G\|_2.
	\vspace{-2mm}
\end{equation}

The input to the discriminator $D$ is the ground truth image $I$ and $I_G$. The GAN loss is as follows: 
\vspace{-2mm}
\begin{equation}
	\mathcal{L}_{GAN}=\log D(I)+\log (1-D(I_G)).
	\vspace{-2mm}
\end{equation}

Furthermore, the perceptual loss  $\mathcal{L}_{per}$ is also adapted to calculate the distance between activation maps of the pre-trained VGG-19 network :

\vspace{-2mm}
\begin{equation}
		\mathcal{L}_{per}=\sum_{i}\left\|\phi_{i}(\tilde{I}_{g})-\phi_{i}\left(I\right)\right\|_{1},
		\vspace{-2mm}
\end{equation}
where $\phi_{i}$ is the activation map of the i-th layer of the VGG-19 network~\cite{simonyan2014very}. During training, each module in OPT are trained separately utilizing corresponding loss combination.

\begin{figure}[t]
	\centering
	\includegraphics[width=\columnwidth]{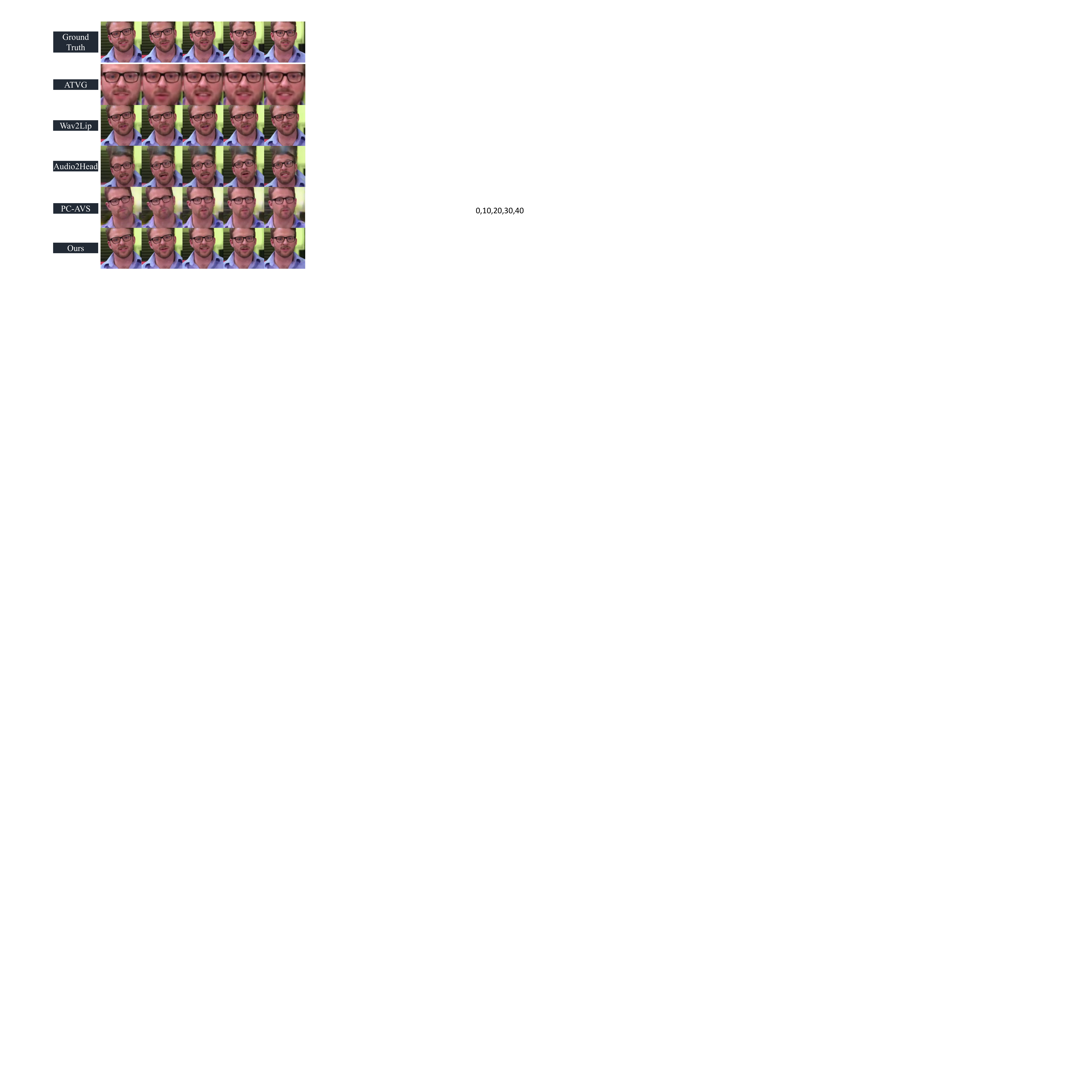} 
	\vspace{-6mm}
	\caption{{Qualitative comparison results with other state-of-the-art methods on LRS2 dataset. }   }
	\label{fig:quality}
\end{figure}

\section{Experiments}
\subsection{Experimental settings}
\textbf{Datasets} The AFDM is trained on audio-visual dataset MEAD~\cite{wang2020mead}  which contains annotations of speaking corpus scripts and identity information. Other modules of OPT are trained and tested on LRW~\cite{chung2016lip} and LRS2~\cite{afouras2018deep} datasets. The LRW dataset contains over 1000 utterances of 500 different words while the LRS2 dataset includes over 140,000 utterances of different sentences. Both videos are from BBC television in the wild. 

\begin{table}[t]
	\centering
	\resizebox{\columnwidth}{!}{
		\begin{tabular}{c c c c c}
			\toprule
			Methods           & SSIM $\uparrow$  & CSIM $\uparrow$ & LMD $\downarrow$ & LSE-C $\uparrow$ \\
			\midrule
			ATVG              &  0.781           &0.76  &       5.32                & 4.165 \\
			Wav2Lip          &  {0.792}           &0.81  & 5.73                      & \textbf{7.237 } \\
			Audio2Head     &  0.743          & 0.72  &  7.34                     &  2.135 \\
			PC-AVS           &   \underline{0.815}          &0.74  & {6.14}  & 6.420  \\
			\midrule
			Ours-Fix Pose  &    0.795         &\underline{0.83}  &\underline{5.25}                  & {6.432} \\
			OPT (Ours) &  \textbf{0.823}& \textbf{0.88} &  \textbf{3.78} &  \underline{6.619} \\
			\bottomrule
		\end{tabular}
	}
		\vspace{-3mm}
	\caption{Quantitative comparison results on LRW dataset. The \textbf{bold} and \underline{underlined} indicate the top-2 results. }
	\vspace{-1mm}
	\label{tab:quantity}
\end{table}

\noindent \textbf{Implementation Details} Face frames are cropped to $256\times256$ size at 25 FPS  and audio to mel-spectrogram of size $16 \times 16$ per frame. Mel-spectrograms are constructed  from 16kHZ audio, window size 800, and hop size 200.  Both encoders in the AFDM share the SE-ResNet~\cite{hu2018squeeze} architecture. $F_{src}$ is extracted by pre-trained ArcFace~\cite{deng2019arcface} model. OPT is trained in stages on 4 Tesla 32G V100 GPUs. The ADAM optimizer is adopted with an initial leaning rage as $10^{-4}$. The learning rate is decreased to $2 \times 10^{-5}$ after $300k$ iterations.

\noindent \textbf{Comparing Methods} The following SOTA one shot talking head generation methods are compared.
\textbf{ATVG}~\cite{chen2019hierarchical} generates frames based on facial landmarks using the attention mechanism.
 \textbf{Wav2Lip}~\cite{prajwal2020lip} utilizes a pre-trained lip-sync discriminator to focus on editing the mouth shape.
\textbf{Audio2Head}~\cite{wang2021audio2head} infers unique head pose sequences from audio and utilizes flow-based generator to produce talking heads.
\textbf{PC-AVS}~\cite{zhou2021pose} extracts modularized audio-visual representations of identity, pose and speech content, generating pose-controllable talking heads. Besides, when $I_{pose}$ is not given for OPT, we can fix it with the same pose as $I_{src}$, keeping the head still. We refer results under this setting as \textbf{Ours-Fix Pose}.  Same as the generation paradigm in PC-AVS~\cite{zhou2021pose},  an extra video with supposedly the same pose as driving frames corresponding to $I_{dri}$ but different identities and mouth shapes is generated to serve as the $I_{pose}$ in our method.

\vspace{-2mm}
\subsection{Quantitative Results}
We evaluate the performance on image quality,  identity preserving and lip-sync quality. The SSIM~\cite{wang2004image} scores are utilized to judge the talking head image quality. CSIM indicates the cosine similarity between face recognition features~\cite{deng2019arcface} of generated and ground truth talking heads.
For the lip-sync quality, the Landmark Distance(LMD) and Lip-Sync Error-Confidence(LSE-C)~\cite{prajwal2020lip} are applied. 

Table \ref{tab:quantity} shows the quantitative comparison results using LRW datasets.  It shows that OPT achieves leading SSIM, LMD and CSIM scores.  As mentioned in~\cite{zhou2021pose}, the leading LSE-C score only means that Wav2Lip is comparable to the ground truth, not better. Our leading LMD score indicates that we preserve better facial structure and accurate mouth shape. We also achieve high-level identity preserving quality, proved by the leading CSIM score.

\begin{figure}[t]
	\centering
	\includegraphics[width=\columnwidth]{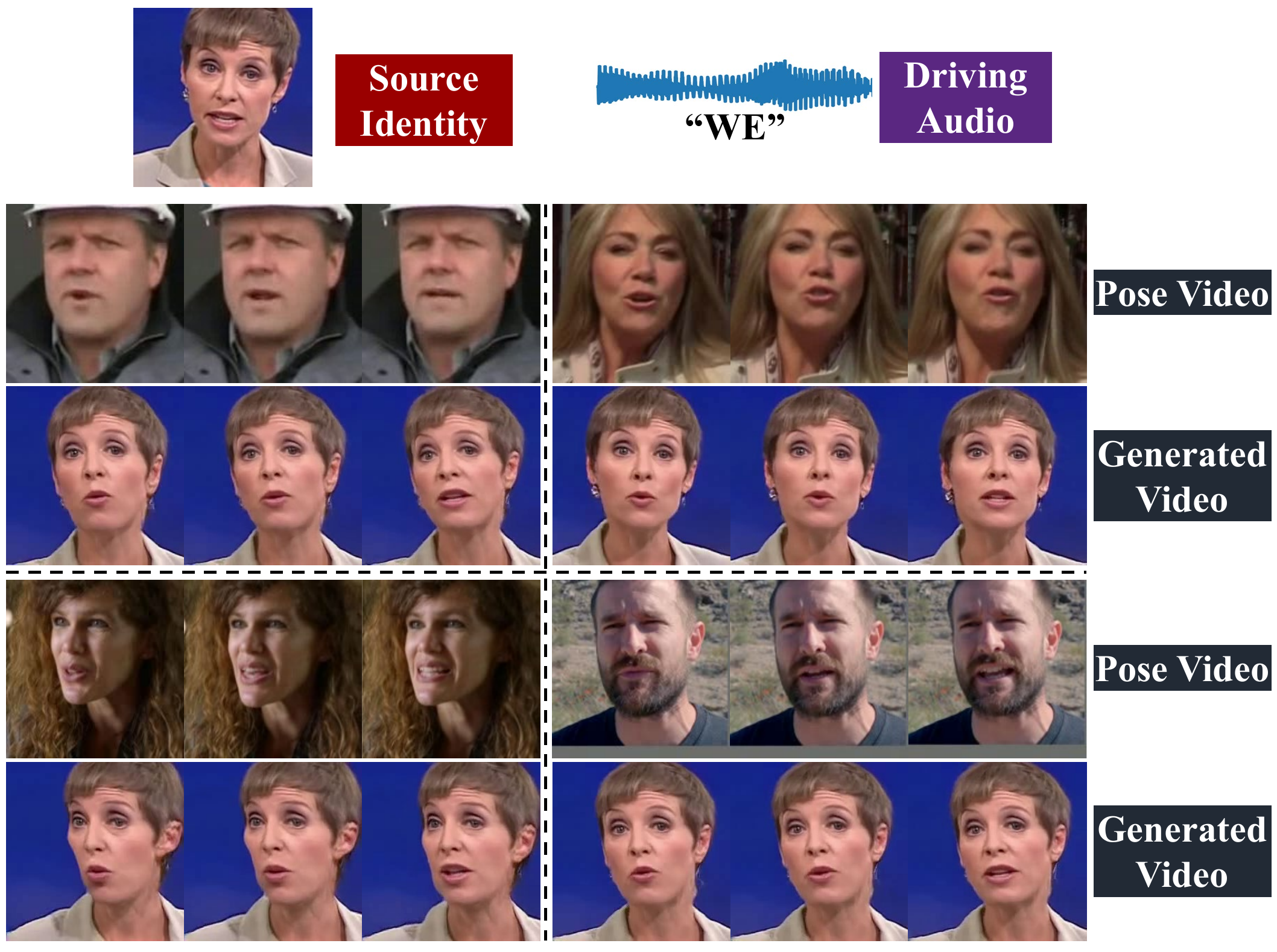} 
		\vspace{-8mm}
	\caption{{Qualitative results of OPT. }  Four generated clips of the same word "we" driven by different pose videos are shown.  }
	\label{fig:quality_pose}
\end{figure}

	\vspace{-3mm}
\begin{table}[t]
	\centering
	\resizebox{\columnwidth}{!}{
		\begin{tabular}{c c c c c}
			\toprule
			Methods           & SSIM $\uparrow$  & CSIM $\uparrow$ & LMD $\downarrow$ & LSE-C $\uparrow$ \\
			\midrule
			Baseline   &    0.721         & 0.69 &       7.18                &  3.125  \\
			Ours w/o AFDM             &    0.753         & 0.75 &         5.78              & 4.259  \\
			
			Ours w/o $\mathcal{L}_{l d m k}$  &    0.762       &  0.74&        6.14            & 3.842  \\
			Ours &  \textbf{0.823}& \textbf{0.88} &  \textbf{3.78} &  \textbf{6.619} \\
			\bottomrule
		\end{tabular}
	}
		\vspace{-4mm}
	\caption{{Ablation study on LRW dataset.} The evaluated parts include the \textit{Audio Feature Disentanglement Module} and the landmark loss utilized in the Audio-to-Expression Module.}
	\vspace{-2mm}
	\label{tab:ablation}
\end{table}

\subsection{Qualitative Results}
We compare OPT with other methods, as displayed in Fig. \ref{fig:quality}. It shows that  OPT generates high quality talking head videos with accurate mouth shape and facial contour that best match the ground truth. Concretely, ATVG merely focuses on cropped facial region. Wav2Lip generates faces with fixed head motions and blurry mouth shape. Audio2Head fails to keep the lip synchronization and accurate facial contour. PC-AVS causes the identity mismatch problem and cannot preserve the accurate facial contour. We further demonstrate the performance of OPT in Fig. \ref{fig:quality_pose}. It indicates that the generated talking heads can achieve free pose control and meanwhile maintain accurate lip synchronization and no identity mismatch problem.

\vspace{-2mm}
\subsection{Ablation Study}
Table \ref{tab:ablation} shows our ablation study to prove the effect of each proposed component. The baseline method directly extracts expression feature from audio signal and utilizes merely reconstruction loss in the Audio Feature Disentanglement Module and Audio-to-Expression Module. The result indicates that AFDM helps solve the identity mismatch problem according to the obvious increase on CSIM score. Besides, $\mathcal{L}_{l d m k}$ contributes a lot to the image quality and lip-sync quality since it extracts accurate facial structure representation during the training process.

\vspace{-2mm}
\subsection{User Study}
We further conduct the user study to evaluate OPT and other state-of-the-art methods. For OPT and comparing methods, 15 video clips using randomly selected source faces and audios from LRW and LRS2 datasets are generated.  We adopt the widely used Mean Opinion Scores (MOS) rating protocol. 10 participants are required to give their ratings (1-5) on the following three aspects for each generated talking head video: visual quality, lip-sync quality and identity preserving quality. As Table \ref{tab:user_study} shows, OPT achieves the best results on all aspects, especially the  identity preserving quality.

\vspace{-2mm}
\section{Conclusion}
In this paper, we propose a new method called OPT to generate pose-controllable identity-preserving talking head videos. Given one source face image and arbitrary driving audio, OPT generates lip-sync talking heads that preserve the source identity and can achieve pose control guided by auxiliary pose video. The proposed Audio Feature Disentanglement Module separates content features from audio signal and the landmark loss is adopted in the Audio-to-Expression Module, both contributing a lot to the generation of talking heads. In the future, we will focus on real time and high resolution generation to enhance the generalization.

\begin{table}[t]
	\centering
	\begin{tabular}{c c c c}
		\toprule
		Method & Visual  & Lip-Sync  & Identity \\
		\midrule
		ATVG          &  2.39  &3.69    &2.81     \\
		Wav2Lip       &  3.57  &4.10 & 3.92   \\
		Audio2Head    &3.31  & 2.45 &  3.63 \\
		PC-AVS        &  3.92&3.87  & 3.14    \\
		Ours          &  \textbf{4.16}& \textbf{4.31} &  \textbf{4.27} \\
		\bottomrule
	\end{tabular}
		\vspace{-2mm}
	\caption{{User study results by mean opinion scores.}}
	\vspace{-3mm}
	\label{tab:user_study}
\end{table}

\bibliographystyle{IEEEbib}
\bibliography{icassp23}

\end{document}